\documentclass[runningheads]{llncs}
\usepackage{graphicx}
\usepackage{comment}
\usepackage{amsmath,amssymb} 
\usepackage{color}

\usepackage{ifthen}
\usepackage{booktabs}
\usepackage{multirow}
\usepackage{colortbl}
\usepackage[table]{xcolor}
\usepackage[pdftex, pagebackref, colorlinks, citecolor=blue, linkcolor=blue]{hyperref}

\begin{document}

\newboolean{revising}
\setboolean{revising}{true}

\ifthenelse{\boolean{revising}} {
\newcommand{\add}[1]{\textcolor{green}{#1}}
\newcommand{\delete}[1]{\textcolor{red}{\sout{#1}}}
\newcommand{\remove}[1]{\delete{#1}}
\newcommand{\replace}[2]{\textcolor{red}{\sout{#1}} \textcolor{green}{#2}}
\newcommand{\unsure}[1]{\textcolor{darkgreen}{#1}}
\newcommand{\revise}[1]{\textcolor{blue}{#1}}
\newcommand{\todo}[1]{\textcolor{red}{TODO: #1}}
\newcommand{\yylin}[1]{\textcolor{blue}{yylin: #1}}
\newcommand{\yht}[1]{\textcolor{blue}{yht: #1}}
\newcommand{\cchsu}[1]{\textcolor{magenta}{cchsu: #1}}
}
{
\newcommand{\revise}[1]{}
\newcommand{\add}[1]{#1}
\newcommand{\delete}[1]{}
\newcommand{\replace}[2]{#2}
\newcommand{\unsure}[1]{}
\newcommand{\remove}[1]{}
\newcommand{\todo}[1]{}
\newcommand{\yylin}[1]{}
\newcommand{\cchsu}[1]{}
}



\def\eg{\emph{e.g.}} \def\Eg{\emph{E.g.}}
\def\ie{\emph{i.e.}} \def\Ie{\emph{I.e.}}
\def\cf{\emph{c.f.}} \def\Cf{\emph{C.f.}}
\def\etc{\emph{etc.}} \def\vs{\emph{vs.}}
\def\wrt{w.r.t.} \def\dof{d.o.f.}
\def\etal{\emph{et al.}}

\newcommand{\bw}{\mathbf{w}}
\newcommand{\bc}{\mathbf{c}}
\newcommand{\bL}{\mathbf{L}}
\newcommand{\bx}{\mathbf{x}}
\newcommand{\bbR}{\mathbb{R}}
\newcommand{\mcalS}{\mathcal{S}}
\newcommand{\mcalB}{\mathcal{B}}
\newcommand{\mcalC}{\mathcal{C}}
\newcommand{\delx}{\Delta_{x}}
\newcommand{\dely}{\Delta_{y}}
\newcommand{\delw}{\Delta_{w}}
\newcommand{\delh}{\Delta_{h}}
\newcommand{\hatx}{\hat{x}}
\newcommand{\haty}{\hat{y}}
\newcommand{\hatw}{\hat{w}}
\newcommand{\hath}{\hat{h}}
\newcommand{\hatmcalB}{\mathcal{B}}
\newcommand{\dxdy}{\mathrm{d} x \mathrm{d} y}

\newcommand{\First}[1]{\textbf{#1}}

\newcommand{\greyrule}{\arrayrulecolor{black!30}\midrule\arrayrulecolor{black}}

\pagestyle{headings}
\mainmatter
\def\ECCVSubNumber{948}  

\title{Every Pixel Matters: Center-aware Feature Alignment for Domain Adaptive Object Detector} 

\titlerunning{Center-aware Feature Alignment for Domain Adaptive Object Detector}
%
\author{
    Cheng-Chun Hsu\inst{1} \and
    Yi-Hsuan Tsai\inst{2} \and
    Yen-Yu Lin\inst{1,3} \and
    Ming-Hsuan Yang\inst{4,5}
}
\authorrunning{C. Hsu et al.}
%
\institute{
    $^1$Academia Sinica\hspace{0.5in}
    $^2$NEC Labs America \\
    $^3$National Chiao Tung University \hspace{0.3in}
    $^4$UC Merced \hspace{0.3in}
    $^5$Google Research \\
}
\maketitle

\begin{abstract}
A domain adaptive object detector aims to adapt itself to unseen domains that may contain variations of object appearance, viewpoints or backgrounds.
Most existing methods adopt feature alignment either on the image level or instance level.
However, image-level alignment on global features may tangle foreground/background pixels at the same time, while instance-level alignment using proposals may suffer from the background noise.
Different from existing solutions, we propose a domain adaptation framework that accounts for each pixel via predicting pixel-wise objectness and centerness.
Specifically, the proposed method carries out center-aware alignment by paying more attention to foreground pixels, hence achieving better adaptation across domains.
%
We demonstrate our method on numerous adaptation settings with extensive experimental results and show favorable performance against existing state-of-the-art algorithms.
Source codes and models are available at \url{https://github.com/chengchunhsu/EveryPixelMatters}.
\end{abstract}


 \begin{figure}[t]
	\centering
    \resizebox{1.0\columnwidth}{!}{
	   \includegraphics{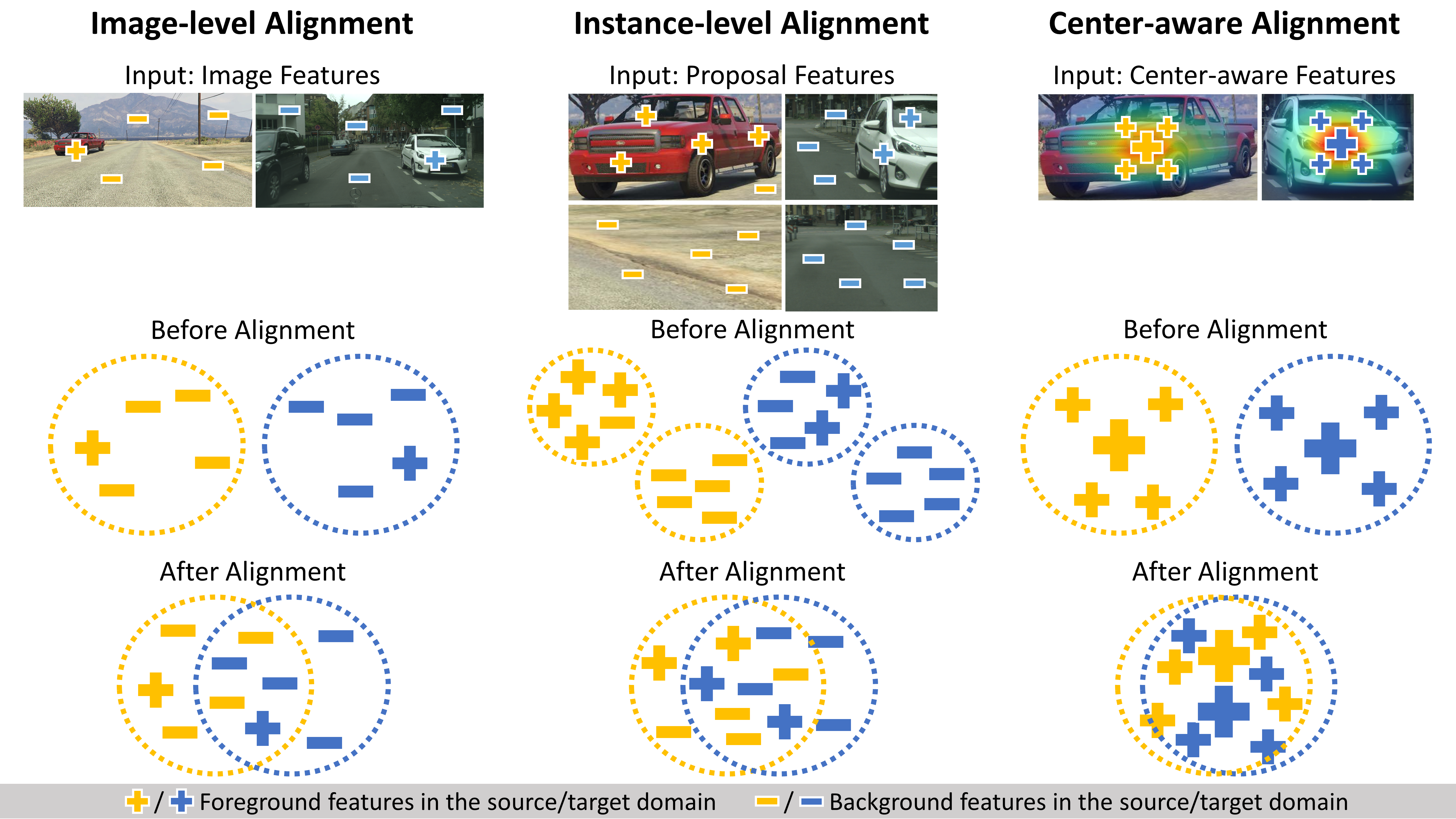}
	}
	\caption{
    	\textbf{Comparisons between different alignment methods.}
    	1) For image-level alignment, it considers both foreground/background pixels, which may lead to noisy alignment and focus more on background pixels.
    	2) Instance-level alignment is performed on proposals, in which the pooled feature on all the pixels within the proposal could mix foreground/background signals. In addition, proposals in the target domain may contain much more background pixels due to the domain gap.
    	3) The proposed center-aware alignment focuses on foreground pixels with higher confidence scores of objectness and centerness, i.e., those marked by larger ``+'' showing higher centerness response, which play a crucial role to reduce the confusion during alignment.
	}
	\label{fig:teaser}
\end{figure}

\section{Introduction}
\label{Introduction}



%
As a key component to image analysis and scene understanding, object detection is essential to many high-level vision applications such as instance segmentation~\cite{DaiCVPR16,HariharanECCV14,HariharanCVPR15,KHeICCV17}, image captioning~\cite{KXuICML15,KarpathyCVPR15,QWuTPAMI18}, and object tracking~\cite{KKangTCSVT18}.
%
Although significant progress on object detection~\cite{GirshickICCV15,RedmonCVPR16,WLiuECCV16} had been made,
an object detector that can adapt itself to variations of object appearance, viewpoints, and backgrounds~\cite{YChenCVPR18} is always in demand.
For example, a detector used for autonomous driving is required to work well under diverse weather conditions, even if training data may be acquired under some particular weather conditions. 
%


%
To address this challenge, \textit{unsupervised domain adaptation} (UDA) methods \cite{LongICML15,GaninICML15,tzeng2017adversarial,Saito_CVPR_2018,Tsai_DA4Seg_ICCV19} have been developed to adapt models trained on an annotated source domain to another unlabeled target domain.
%
Adopting a similar strategy to the classification task \cite{tzeng2017adversarial} using adversarial feature alignment, numerous UDA methods for objection detection~\cite{ZHeICCV19,SaitoCVPR19,InoueCVPR18,QCaiCVPR19,KimICCV19,KimCVPR19,YChenCVPR18,Hsu_WACV_2020} are proposed to reduce the domain gap across source and target domains.
%
%
However, such alignment is usually performed on the image level that adapts global features, which is less effective when the domain gap is large \cite{SaitoCVPR19,dai2019adaptation}.
To improve upon global alignment\footnote{In this paper, we use image-level alignment and global alignment interchangeably.}, existing methods \cite{YChenCVPR18,ZHeICCV19,XZhuCVPR19} adapt instance-level distributions that pool features of all the pixels within a proposal.
However, since pixel distributions are unknown in the target domain, the proposal extracted from the target domain could contain many background pixels.
As a result, this may significantly confuse the alignment procedure when adapting instance-level features of target proposals to the source distribution that contains mostly foreground pixels (Fig. \ref{fig:teaser}).

In this paper, we propose to take every pixel into consideration when aligning feature distributions across two domains.
To this end, we design a module to estimate pixel-wise objectness and centerness of the entire image, which allows our alignment process to focus on foreground pixels, instead of the proposal that may contain tangled foreground/background pixels as considered in the prior work. 
In order to predict the pixel-wise information, we revisit the object detection framework and adopt fully-convolutional layers.
%
As a result, our method aims to align the centered discriminative part of the objects across domains, namely the regions with high objectness scores and close to the object centers (see Fig. \ref{fig:teaser}).
%
%
%
%
Thereby, these regions are less sensitive to irrelevant background pixels in the target domain and facilitate distribution alignment.
To the best of our knowledge, we make the first attempt to leverage pixel-wise objectness and centerness for domain adaptive object detection.

To validate the proposed method, we conduct extensive experiments on three benchmark settings for domain adaptation: Cityscapes \cite{Cordts2016Cityscapes} $\rightarrow$ Cityscapes Foggy \cite{SDV18}, Sim10k \cite{sim10k} $\rightarrow$ Cityscapes, and KITTI \cite{Geiger2012CVPR} $\rightarrow$ Cityscapes.
The experimental results show that our center-aware feature alignment performs favorably against existing state-of-the-art algorithms.
%
Furthermore, we provide ablation study to demonstrate the usefulness of each component in our method.
The major contributions of this paper are summarized as follows.
First, we propose to discover discriminative object parts on the pixel level and better handle the domain adaptation task for object detection.
%
%
Second, center-aware distribution alignment with its multi-scale extension is presented to account for object scales and alleviate the unfavorable effects caused by cluttered backgrounds during adaptation.
%
%
%
%
%
Third, comprehensive ablation studies validate the effectiveness of the proposed framework with center-aware feature alignment. 
%
\section{Related Work}
In this section, we review a few research topics relevant to this work, including object detection and domain adaptive object detection.

\subsection{Object Detection}
Object detection studies can be categorized into anchor-based and anchor-free detectors. 
%
Anchor-based detectors compile a set of anchors to generate object proposals, and formulate object detection as a series of classification tasks over the proposals.
Faster-RCNN~\cite{SRenNIPS15} is the pioneering anchor-based detector,
where the region proposal network (RPN) is employed for proposal generation. 
Owing to its effectiveness, RPN is widely adopted in many anchor-based detectors \cite{TLinICCV17,WLiuECCV16}.
Anchor-free detectors skip proposal generation, and directly localize objects based on the fully convolutional network (FCN)~\cite{LongCVPR15}.
%
%
%
Recently, anchor-free methods~\cite{HLawECCV18,XZhouCVPR19,KDuanICCV2019} leverage keypoint (i.e., the center or corners of a box) localization and achieve comparable performance with anchor-based methods. 
Yet, these methods require complex post-processing for grouping the detected points. 
To avoid such a process, FCOS~\cite{ZTianICCV19} proposes per-pixel prediction, and directly predicts the class and offset of the corresponding object at each location on the feature map.
In this work, we take advantages of the property in anchor-free methods to identify discriminate areas for the alignment procedure.

\begin{table*} [t]
	\caption{
		\textbf{Alignment schemes adopted by existing methods}, including global alignment ($G$), instance-level alignment ($I$), low-level feature alignment ($L$), pixel-level alignment ($P$) via style transfer or CycleGAN, pseudo-label re-training (\textit{PL}), and the proposed center-aware alignment (\textit{CA}) that considers pixel-wise objectness and centerness. $^*$ indicates that pixel-level alignment is only applied during adapting from Sim10k to Cityscapes.
	}
	\label{table:alignment_methods}
	\footnotesize
	\centering
	\begin{tabular}{l@{\hskip 15pt}c@{\hskip 15pt}c@{\hskip 15pt}c@{\hskip 15pt}c@{\hskip 15pt}c@{\hskip 15pt}c}
		\toprule
		
        Method & $G$ & $I$ & $L$ & $P$ & \textit{PL} & \textit{CA} \\
		
		\midrule

        DAF~\cite{YChenCVPR18} \textsubscript{CVPR'18}   & ${\surd}$ & ${\surd}$ &      &      &            &           \\
        SC-DA~\cite{XZhuCVPR19} \textsubscript{CVPR'19}  & ${\surd}$ & ${\surd}$ &      &      &            &           \\
        SW-DA~\cite{SaitoCVPR19} \textsubscript{CVPR'19} & ${\surd}$ &           & ${\surd}$ & ${\surd}$* &     &       \\
        DAM~\cite{KimCVPR19} \textsubscript{CVPR'19}     & ${\surd}$ &           &           & ${\surd}$  &     &       \\
        MAF~\cite{ZHeICCV19} \textsubscript{ICCV'19}     & ${\surd}$ & ${\surd}$ &           &            &     &       \\
        MTOR~\cite{QCaiCVPR19} \textsubscript{CVPR'19}   & ${\surd}$ & ${\surd}$ &           &            &      &      \\
        STABR~\cite{KimICCV19} \textsubscript{ICCV'19}   & ${\surd}$ &           &           &            &  ${\surd}$   &       \\
        PDA~\cite{Hsu_WACV_2020} \textsubscript{WACV'20}   & ${\surd}$ &           &           &   ${\surd}$  &    &       \\
        Ours                                             & ${\surd}$ &           &      &      &            & ${\surd}$ \\
        
        \bottomrule
	\end{tabular}
\end{table*}

\subsection{UDA for Object Detector}

Chen~\etal~\cite{YChenCVPR18} first present two alignment practices, \ie, image-level and instance-level alignments, by adopting adversarial learning at image and instance scales, respectively.
For image-level alignment, Saito~\etal~\cite{SaitoCVPR19} further indicate that aligning lower-level features is more effective since global feature alignment suffers from the cross-domain variations of foreground objects and background clutter.
%
%
To improve instance-level alignment, Zhu~\etal~\cite{XZhuCVPR19} apply $k$-means clustering to group proposals and obtain the centroids of these clusters, which achieves a balance between global and instance-level alignment. 
However, their method introduces additional data-independent hyper-parameters for clustering and is not end-to-end trainable. 
Other variants improve feature alignment based on a hierarchical module \cite{ZHeICCV19}, a style-transfer based method to address the source-biased issue \cite{KimCVPR19}, a teacher-student scheme to explore object relations \cite{QCaiCVPR19}, and a progressive alignment scheme \cite{Hsu_WACV_2020}.
%
%
%
%

While the above methods are based on two-stage detectors, Kim~\etal~\cite{KimICCV19} propose a one-stage adaptive detector for faster inference, via a hard negative mining technique for seeking more reliable pseudo-labels.
%
%
However, their method only partially alleviates the issues brought by background and does not consider every pixel during feature alignment to reduce the domain gap.
%
We also note that all aforementioned methods are based on anchors, in which performing instance-level alignment would be sensitive to inaccurate proposals in the target domain and the mixture of foreground/background pixels in a proposal.
In contrast, we address these drawbacks by predicting pixel-wise objectness and proposing center-aware feature alignment, which only focuses on the discriminative parts of objects at the pixel scale.
In Table \ref{table:alignment_methods}, we summarize the alignment methods used in the aforementioned techniques for domain adaptive object detection.
\section{Proposed Method}

\begin{figure*}[t]
	\centering
	    \includegraphics[width=0.9\linewidth]{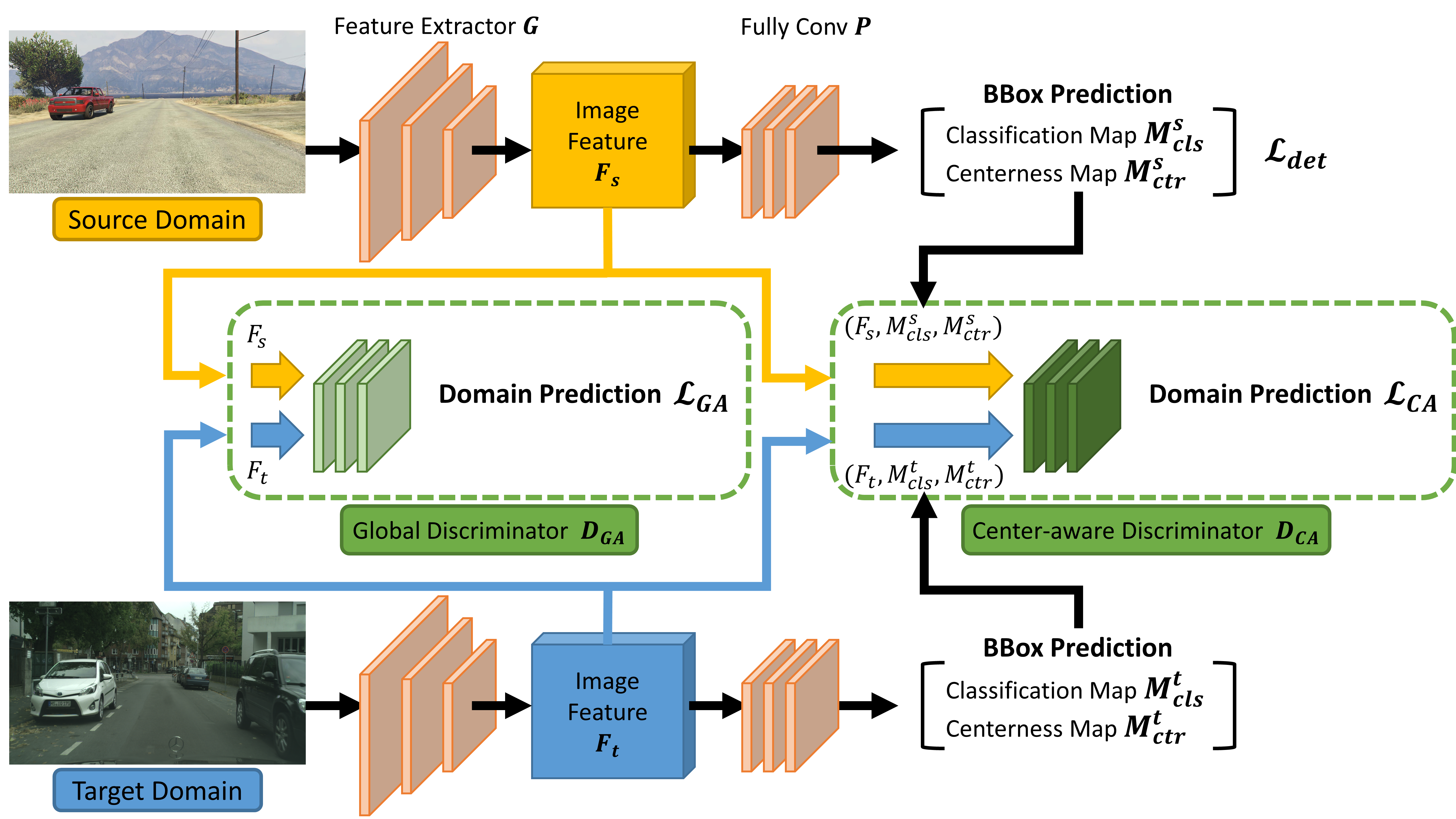}
4

	\caption{
    	\textbf{Proposed framework for domain adaptive object detection}.
    	Given the source and target images, we feed them to a shared feature extractor $G$ to obtain their features $F$. Then, the global alignment on these features is performed via a global discriminator $D_{GA}$ and a domain prediction loss $\mathcal{L}_{GA}$.
    	Next, we pass the feature through the fully-convolutional module $P$ to produce the classification and centerness maps. 
    	These maps and the feature $F$ are utilized to generate the center-aware features.
    	Finally, we use a center-aware discriminator $D_{CA}$ and another domain prediction loss $\mathcal{L}_{CA}$ to perform the proposed center-aware feature alignment.
    	Note that the bounding box prediction loss $\mathcal{L}_{det}$ is only operated on source images using their corresponding ground-truth bounding boxes.
	}
	\label{fig: architecture}
\end{figure*}


In this section, we first describe global feature alignment, and then introduce the proposed center-aware alignment that utilizes pixel-wise objectness and ceterness.
To improve the performance, we further incorporate multi-scale alignment that takes object scale into account during adaptation.

\subsection{Algorithm Overview}

Given a set of source images $I_s$, their ground-truth bounding boxes $B_s$, and unlabeled target images $I_t$, our goal is to predict bounding boxes $B_t$ on the target image.
To this end, we propose to utilize two alignment schemes that complement each other: global alignment that accounts for image-level distributions and the proposed center-aware alignment that focuses more on foreground pixels.
The overall procedure is illustrated in Fig. \ref{fig: architecture}.
Given a shared feature extractor $G$ across domains, we first extract features $F = G(I)$ and perform global alignment via using a global discriminator and a domain prediction loss. Second, followed by $G$, a fully-convolutional module $P$ is adopted to predict pixel-wise objectness and centerness maps. Through combining these maps with the feature $F$, we employ another center-aware discriminator and its domain prediction loss to perform center-aware alignment. 

\subsection{Global Feature Alignment}
\label{GlobalAlignment}
%
%
%
The goal of global alignment is to align the feature maps on the image level to reduce the domain gap. To this end, we apply the adversarial alignment technique \cite{YChenCVPR18} via utilizing a global discriminator~${D}_{GA}$, which aims to identify whether the pixels on each feature map come from the source or the target domain.
%

Particularly, given the $K$-dimensional feature map~$F \in \bbR^{H \times W \times K}$ of the spatial resolution $H \times W$ from the feature extractor $G$, the output of ${D}_{GA}$ is a domain classification map that has the same size as $F$, while each location represents the domain label corresponding to the same location on $F$.
Note that we set the domain label $z$ of source and target domain as $1$ and $0$, respectively. Therefore, the discriminator can be optimized by minimizing the binary cross-entropy loss. For a location $(u, v)$ on $F$, the loss function can be written as
\begin{equation}
\label{eqn:loss_ga}
\mathcal{L}_{GA}(I_s, I_t) = - \sum_{u, v} z \: \log({D}_{GA}(F_s)^{(u,v)}) + (1-z) \log(1-{D}_{GA}(F_t)^{(u,v)}).
\end{equation} 

To perform adversarial alignment, we apply the gradient reversal layer~(GRL) \cite{GaninICML15} to feature maps of both source/target images, in which the sign of the gradient is reversed when optimizing the feature extractor via the GRL layer.
%
%
Then the mechanism works as follows. The loss for the discriminator is minimized via \eqref{eqn:loss_ga}, while the feature extractor is optimized by maximizing this loss, in order to deceive the discriminator.
We also note that most existing methods (those in Table \ref{table:alignment_methods}) utilize such global alignment that focuses on image-level distributions (i.e., more background pixels in reality). We also use global alignment in our framework to complement the proposed center-aware alignment that focuses on foreground pixels. 
%

\begin{figure}[t]
\centering
\resizebox{0.8\columnwidth}{!}{
\includegraphics{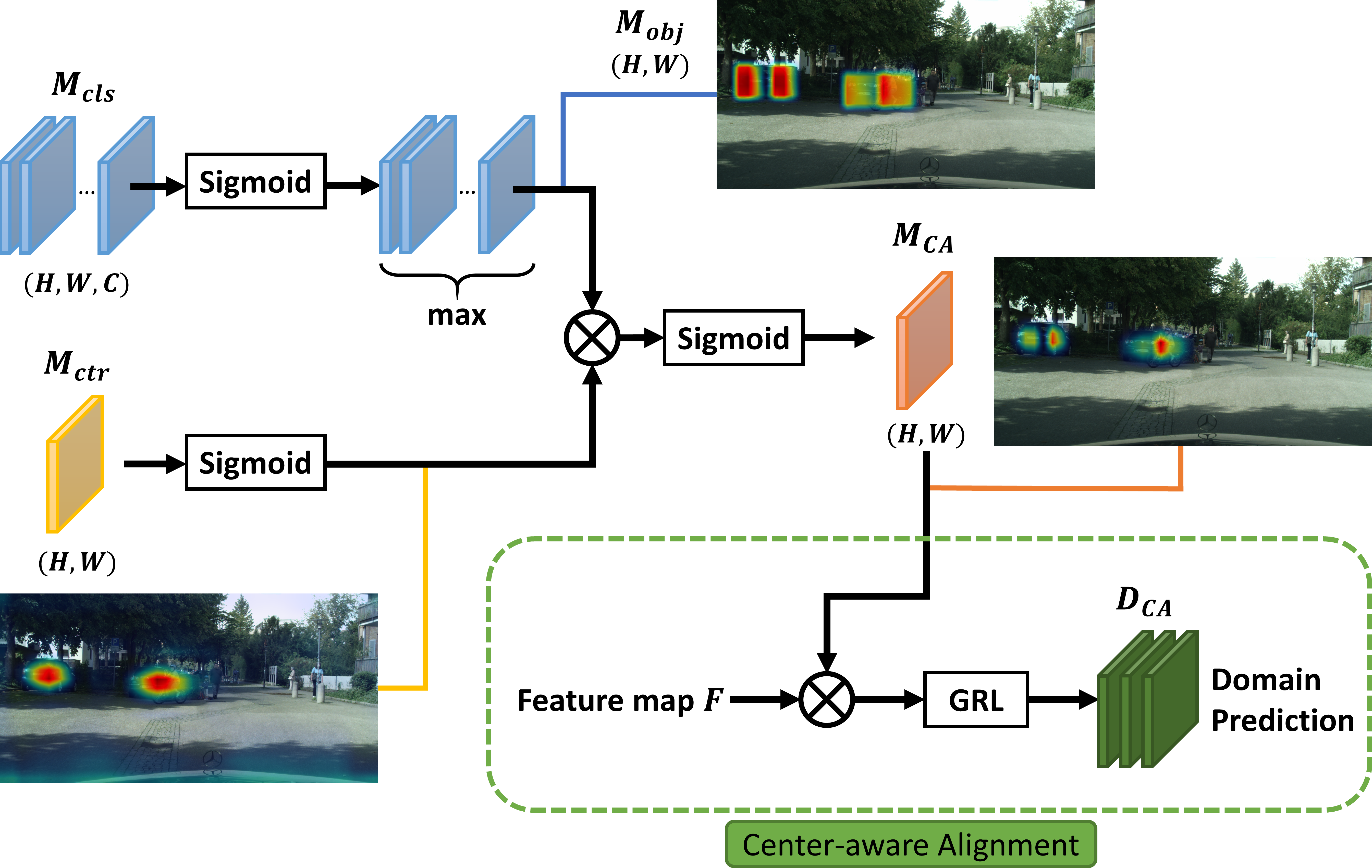}
}

\caption{
\textbf{Proposed center-aware alignment.}
Given the classification output $M_{cls}$, we first convert it to a class-agnostic map $M_{obj}$, which is then merged with the centerness output $M_{ctr}$ into a center-aware map $M_{CA}$ via \eqref{eqn:eq_obj_map} to identify potential object locations.
Next, we use this map $M_{CA}$ as the guidance to weight the global feature map $F$.
Finally, this weighted feature map serves as the input to the center-aware discriminator $D_{CA}$ to enable the proposed center-aware alignment in the feature space via \eqref{eqn:loss_ca}.
}
\label{fig: CA_alignment}
\end{figure}

\subsection{Center-aware Alignment}
\label{Center-awareAlignment}
As mentioned in Section~\ref{Introduction} and Table \ref{table:alignment_methods}, existing methods \cite{YChenCVPR18,ZHeICCV19,XZhuCVPR19} for instance-level alignment are based on proposals, and thus these approaches may suffer from the background effect.
%
In order to address this issue, we propose a center-aware alignment method that allows us to focus on discriminative object regions.
To this end, we adopt a center-aware discriminator ${D}_{CA}$ for aligning features in the high-confidence area on the pixel level.

{\flushleft \bf{Definition.}}
With a designed fully-convolutional network $P$ (as detailed in Section~\ref{ArchitectureOptimization}) and feature map~$F \in \bbR^{H \times W \times K}$ from the feature extractor $G$, we pass $F$ through $P$, and obtain a classification output $M_{cls} \in \bbR^{H \times W \times C}$ and a class-agnostic centerness output $M_{ctr} \in \bbR^{H \times W}$, where $C$ is the number of categories. Each location on the classification and centerness maps indicates corresponding objectness and centerness scores, respectively.

%
%
{\flushleft \bf{Discover Object Region.}}
In order to find the confident area containing foreground objects, we utilize two cues derived from our object detector as mentioned above: 1) a class-agnostic map of the objectness scores and 2) a centerness map that highlight object centers, so that the alignment can focus more on object parts.
First, the objectness map can be obtained from the classification output $M_{cls}$.
%
To obtain the class-agnostic map, we apply the \textit{sigmoid} activation on each channel and take the \textit{max} operation over categories.
Similarly, the final class-agnostic centerness map is obtained via applying the \textit{sigmoid} activation on the centerness output $M_{ctr}$.
%
Overall, the final map $M_{CA}$ to guide our center-aware alignment is calculated as follows:
%
\begin{align}
\label{eqn:eq_obj_map}
M_{obj} & = \underset{c}{\operatorname{max}}(\sigma(M_{cls})), \notag \\
M_{CA} & = \sigma(\delta \: M_{obj} \odot \sigma(M_{ctr})),
\end{align} 
where $\sigma$ represents the \textit{sigmoid} activation and $\odot$ denotes the element-wise product, i.e., Hadamard product, on the spatial maps.
Since the values in $M_{obj}$ and $\sigma(M_{ctr})$ are ranged from $0$ to $1$, a scaling factor $\delta$ is introduced for preventing the value from being too small after the multiplication. The factor $\delta$ is set to $20$ in all experiments.

{\flushleft \bf{Perform Alignment.}}
With the center-aware map $M_{CA}$, we are able to highlight the area where alignment on the pixel level should pay attention. To use this map as the guidance to our center-aware alignment, we multiply it by the feature map $F$ and then feed it into the center-aware discriminator ${D}_{CA}$:
\begin{align}
\label{eqn:loss_ca}
\mathcal{L}_{CA}(I_s, I_t) & = -\sum_{u, v} z \: \log({D}_{CA}(M_{CA}^s \odot F_s)^{(u,v)}) \notag\\
                        & + (1-z) \log(1-{D}_{CA}(M_{CA}^t \odot F_t)^{(u,v)}).
\end{align} 
We note that, since $M_{CA}$ is a map of resolution $H \times W$, we duplicate it for $K$ channels to compute its element-wise product with the feature map $F \in \bbR^{H \times W \times K}$.
Then, we adopt a similar alignment process as described in \eqref{eqn:loss_ga} via the GRL layer.
As a result, different from the global alignment method as described in Section \ref{GlobalAlignment}, our model aligns pixel-wise features that are likely to be the object and hence mitigates the non-matching issue between foregrounds and backgrounds. The entire process of center-aware alignment is illustrated in Fig.~\ref{fig: CA_alignment}.

\subsection{Overall Objective for Proposed Framework}
\label{OverallObjective}
Given source images $I_s$, target image $I_t$, and the ground-truth bounding boxes $B_{s}$ in the source domain, our goal is to predict bounding boxes $B_{t}$ on the unlabeled target data.
We have described the objective for feature alignment on both source and target images. Here, we introduce the details of the object detection objective on the source domain using $I_s$ and $B_s$.

{\flushleft \bf{Objective for Object Detector.}}
Motivated by the anchor-free detector \cite{ZTianICCV19}, our fully-convolutional module $P$ consists of the classification, centerness, and regression branches. The three branches output the objectness map $M_{obj}$, centerness map $M_{ctr}$, and regression map $M_{reg}$, respectively. For the classification and regression branches, their goals are to predict the classification score and the distance to the four sides of the corresponding object box for each pixel, respectively.
We denote their loss functions as $\mathcal{L}_{cls}$ and $\mathcal{L}_{reg}$, which can be optimized via the focal loss~\cite{TLinICCV17} and IoU loss~\cite{JYuACMMM16}, respectively.
For the centerness branch, it predicts the distance between each pixel and the center of the corresponding object box and can be optimized by the binary cross-entropy loss \cite{ZTianICCV19} denoted as $\mathcal{L}_{ctr}$. 
The overall objective for the detector on the source domain is:
\begin{equation}
\mathcal{L}_{det}(I_s, B_s) = \mathcal{L}_{cls} + \mathcal{L}_{reg} + \mathcal{L}_{ctr}.
\label{eqn:loss_det}
\end{equation} 
Here, we omit the argument $(I_s, B_s)$ of each loss function for simplicity.

{\flushleft \bf{Overall Objective.}}
In order to obtain domain-invariant features across the source and target domains, we apply adversarial learning to feature maps using two discriminators, ${D}_{GA}$ and ${D}_{CA}$, which perform the global alignment and center-aware alignment by minimizing the objective functions $\mathcal{L}_{GA}$ and $\mathcal{L}_{CA}$, respectively. The details can be found in Section~\ref{GlobalAlignment} and Section~\ref{Center-awareAlignment}.
The overall loss function can be expressed as:
\begin{align}
\mathcal{L}(I_s, I_t, B_s) = \mathcal{L}_{det}(I_s, B_s) + \alpha \mathcal{L}_{GA}(I_s, I_t)
+ \beta \mathcal{L}_{CA}(I_s, I_t), 
\label{eqn:loss_overall}
\end{align} 
where $\alpha$ and $\beta$ are the weights used to balance the three terms. 

\subsection{Network Architecture and Discussions}
\label{ArchitectureOptimization}
Different from the prior work \cite{YChenCVPR18,ZHeICCV19,XZhuCVPR19} that focuses on instance-level alignment, our center-aware feature alignment requires pixel-wise predictions for objectness and centerness maps, so we cannot directly adopt the network architecture in previous methods.
In this section, we introduce our architecture via using a fully-convolutional module for producing pixel-wise predictions, as well as a multi-scale extension to account for the object scale during adaptation.
%
%
%

%
{\flushleft \bf{Network Architecture.}}
As mentioned in Section~\ref{OverallObjective}, we connect feature map $F$ with the fully-convolutional detection head $P$ that contains three branches: the classification, centerness, and regression branches.
Different from previous methods, all branches are constructed by the fully-convolutional network, so that the predictions are performed on the pixel level. Specifically, the three branches consist of four $3 \times 3$ convolutional layers, and each of them has $256$ filters. For both discriminators in global and center-aware alignments, i.e., ${D}_{GA}$ and ${D}_{CA}$, we use the same fully-convolutional architecture as the detection branch, in order to maintain the consistency of the output size and thus map to the original input image.

{\flushleft \bf{Multi-scale Alignment.}}
We observe that such a fully-convolutional architecture is not robust to the object scale, which is crucial to the performance of feature alignment.
Therefore, in the feature extractor $G$, we use the feature pyramid network (FPN)~\cite{TLinCVPR17} to handle different sizes of objects.
%
Particularly, FPN utilizes five levels of feature map, which can be denoted as $F^i$ for $i = \{3,4, ..., 7\}$.
The feature map $F^3$ is responsible for the smallest objects, while the feature map $F^7$ focuses on the largest objects.
Each of the feature maps in the pyramid, i.e., $F^i$, has 256 channels.

We connect each layer with one head that contains three detection branches and two discriminators, i.e., ${D}_{GA}$ and ${D}_{CA}$, and thus the loss function in \eqref{eqn:loss_overall} can be extended to the feature map of each layer.
%
%
As a result, we are able to align each individual feature map $F^i$ via global and center-aware alignments via \eqref{eqn:loss_ga} and \eqref{eqn:loss_ca}.
It follows that each aligned layer is responsible for a certain range of object size while making the overall alignment process consistent.
%


{\flushleft \bf{How Pixel-wise Prediction Helps Feature Alignment.}}
It is worth mentioning that we take advantage of the pixel-wise prediction for the following reasons: 1) Pixel-wise prediction does not involve any fixed anchor-related hyperparameters to produce proposals, which could be biased to the source domain during training;
%
2) Pixel-wise prediction considers all the pixels during training, which helps increase the capability of the model to identify the discriminative area of target objects;
%
3) The alignment can be performed on the pixel level and focuses on foreground pixels, which enables the model to learn better feature alignment.
%
Note that the proposed method only depends on pixel-wise prediction, in which our method can be also applied to other similar detection models using the fully-convolutional module.
		
		
		
        

		
		
		
        

\section{Experimental Results}
We first provide the implementation details, and then describe datasets and evaluation metrics. Next, we compare our method with the state-of-the-art methods on multiple benchmarks. Finally, we conduct further analysis to understand the effect of each component in our framework.
All the source code and models will be made available to the public.

\subsection{Implementation Details}
%

We implement our method with the PyTorch framework. In all the experiments, we set $\alpha$ and $\beta$ in \eqref{eqn:loss_overall} as $0.01$ and $0.1$, respectively.
Considering that center-aware alignment involves the detection output from \eqref{eqn:eq_obj_map}, we first pre-train the detector only with the global alignment as a warm-up stage to ensure the reliability of detection before applying center-aware alignment and training the full objective in \eqref{eqn:loss_overall}.
Note that we set a larger $\alpha$ as $0.1$ during pre-training for a faster convergence.
For the adversarial loss using reversed gradients via GRL, we set the weight as $0.01$ and $0.02$ for $D_{GA}$ and $D_{CA}$, respectively.
The model is trained with learning rate of $5$ $\times$ $10^{-3}$, momentum of $0.9$, and weight decay of $5$ $\times$ $10^{-4}$. 
The input images are resized with their shorter side as 800 and longer side less or equal to 1333.
%

\subsection{Datasets}
We follow the dataset setting as described in \cite{YChenCVPR18} and perform experiments for weather, synthetic-to-real and cross-camera adaptations on road-scene images.

{\flushleft \bf{Weather Adaptation.}}
Cityscapes~\cite{Cordts2016Cityscapes} is a scene dataset for driving scenarios, which are collected in dry weather. It consists of 2975 and 500 images in the training and validation set, respectively. The segmentation mask is provided for each image, consisting of eight categories: \textit{person}, \textit{rider}, \textit{car}, \textit{truck}, \textit{bus}, \textit{train}, \textit{motorcycle} and \textit{bicycle}. The Foggy Cityscapes~\cite{SDV18} dataset is synthesized from Cityscapes as foggy weather. In the experiment, we adapt the model from Cityscapes to Foggy Cityscapes for studying the domain shift caused by the weather condition.

{\flushleft \bf{Synthetic-to-real.}}
Sim10k~\cite{sim10k} is a collection of synthesized images, which consists of 10,000 images and their corresponding bounding box annotations. We use images of Sim10k as the source domain, while Cityscapes is considered as the target domain. The adaptation from Sim10k to Cityscapes is used to evaluate the adaptation ability from synthesized to real-world images. Following the literature, only the class \textit{car} is considered.

{\flushleft \bf{Cross-camera Adaptation.}}
KITTI~\cite{Geiger2012CVPR} is similar to Cityscapes as a scene dataset, except that KITTI has a different camera setup. The training set of KITTI consists of 7,481 images. We use the KITTI and Cityscapes as the source domain and target domain respectively, and evaluation the capability of cross-camera adaptation. Following the literature, only the class \textit{car} is considered.


\begin{table*} [t]
	\caption{
		\textbf{Results of adapting Cityscapes to Foggy Cityscapes.} The first and second groups adopt VGG-16 and ResNet-101 as the backbone, respectively.
		Note that results of each class are evaluated in mAP$^r_{0.5}$.
	}
	\label{table:synthia}
	\footnotesize
	\centering
    \resizebox{\textwidth}{!}{
	\begin{tabular}{lcccccccccc}
		\toprule
		
		& \multicolumn{10}{c}{Cityscapes $\rightarrow$ Foggy Cityscapes} \\
		
		\midrule
        Method & Backbone & person &  rider &  car &  truck &  bus &  train &  mbike &  bicycle &  mAP$^r_{0.5}$ \\
		
		\midrule
        Baseline (F-RCNN)                & \multirow{10}{*}{VGG-16} & 17.8 & 23.6 & 27.1 & 11.9 & 23.8 & 9.1 & 14.4 & 22.8 & 18.8 \\
        DAF~\cite{YChenCVPR18} \textsubscript{CVPR'18}          & & 25.0 & 31.0 & 40.5 & 22.1 & 35.3 & 20.2 & 20.0 & 27.1 & 27.6 \\
        SC-DA~\cite{XZhuCVPR19} \textsubscript{CVPR'19}  & & 33.5 & 38.0 & 48.5 & 26.5 & 39.0 & 23.3 & 28.0 & 33.6 & 33.8 \\
        MAF~\cite{ZHeICCV19} \textsubscript{ICCV'19}             & & 28.2 & 39.5 & 43.9 & 23.8  & 39.9 & 33.3 & \First{29.2} & 33.9 & 34.0 \\
        SW-DA~\cite{SaitoCVPR19} \textsubscript{CVPR'19}       & & 29.9 & \First{42.3} & 43.5 & 24.5 & 36.2 & 32.6 & 30.0 & 35.3 & 34.3 \\
        DAM~\cite{KimCVPR19} \textsubscript{CVPR'19}           & & 30.8 & 40.5 & 44.3 & \First{27.2} & 38.4 & \First{34.5} & 28.4 & 32.2 & 34.6 \\
        
        Ours (w/o adapt.)     & & 30.5 & 23.9 & 34.2 & 5.8  & 11.1 & 5.1  & 10.6 & 26.1 & 18.4 \\
        Ours (GA)            & & 38.7 & 36.1 & 53.1 & 21.9 & 35.4 & 25.7 & 20.6 & 33.9 & 33.2 \\
        Ours (CA)            & & 41.3 & 38.2 & 56.5 & 21.1 & 33.4 & 26.9 & 23.8 & 32.6 & 34.2 \\
        Ours (GA+CA)         & & \First{41.9} & 38.7 & \First{56.7} & 22.6 & \First{41.5} & 26.8 & 24.6 & \First{35.5} & \First{36.0} \\
        
        \greyrule
        Oracle              & & 47.4 & 40.8 & 66.8 & 27.2 & 48.2 & 32.4 & 31.2 & 38.3 & 41.5 \\
        
        \midrule
        
        
        Ours (w/o adapt.)      & \multirow{4}{*}{ResNet-101} & 33.8 & 34.8 & 39.6 & 18.6 & 27.9 & 6.3  & 18.2 & 25.5 & 25.6 \\
        Ours (GA)            & & 39.4 & 41.1 & 54.6 & 23.8 & 42.5 & 31.2 & 25.1 & 35.1 & 36.6 \\
        Ours (CA)            & & 40.4 & \First{44.9} & \First{57.9} & 24.6 & \First{49.6} & 32.1 & 25.2 & 34.3 & 38.6 \\

        Ours (GA+CA)         & & \First{41.5} & 43.6 & 57.1 & \First{29.4} & 44.9 & \First{39.7} & \First{29.0} & \First{36.1} & \First{40.2} \\
        
        \greyrule
        Oracle              & & 44.7 & 43.9 & 64.7 & 31.5 & 48.8 & 44.0 & 31.0 & 36.7 & 43.2 \\
        
        \bottomrule
	\end{tabular}
	}
\end{table*}

\begin{table} [t]
	\caption{
		\textbf{Results of adapting Sim10k/KITTI to Cityscapes.} 
		The first and second groups adopt VGG-16 and ResNet-101 as the backbone, respectively.
		The symbol~$*$ indicates that additional training images generated via pixel-level adaptation are used. 
		%
	}
	\label{table:sim10k_kitti}
	\footnotesize
	\centering
    \resizebox{0.55\textwidth}{!}{
	\begin{tabular}{lcc|c}
		\toprule
		
		\multicolumn{2}{c}{} & \multicolumn{1}{c}{Sim10k} & \multicolumn{1}{c}{KITTI} \\
		
		\midrule
		 Method &  Backbone &  mAP$^r_{0.5}$ & mAP$^r_{0.5}$ \\
		
		\midrule
        Baseline (F-RCNN)                        & \multirow{10}{*}{VGG-16} & 30.1 & 30.2 \\
        DAF~\cite{YChenCVPR18} \textsubscript{CVPR'18}                  & & 39.0 & 38.5 \\
        MAF~\cite{ZHeICCV19} \textsubscript{ICCV'19}                   & & 41.1 & 41.0 \\
        SW-DA~\cite{SaitoCVPR19} \textsubscript{CVPR'19}      & & 42.3 & - \\
        SW-DA*~\cite{SaitoCVPR19} \textsubscript{CVPR'19}    & & 47.7 & - \\
        SC-DA~\cite{XZhuCVPR19} \textsubscript{CVPR'19}         & & 43.0 & 42.5 \\

        Ours (w/o adapt.)      & & 39.8 & 34.4 \\
        Ours (GA)            & & 45.9 & 39.1 \\
        Ours (CA)            & & 46.6 & 41.9 \\
        Ours (GA+CA)         & & \First{49.0} & \First{43.2} \\
        
        \greyrule
        Oracle              & & 69.7 & 69.7 \\
        
        \midrule
        
        
        Ours (w/o adapt.)     & \multirow{4}{*}{ResNet-101} & 41.8 & 35.3 \\
        Ours (GA)            & & 50.6 & 42.3 \\
        Ours (CA)            & & 51.1 & 43.6 \\
        Ours (GA+CA)         & & \First{51.2} & \First{45.0} \\
        
        \greyrule
        Oracle              & & 70.4 & 70.4 \\

        \bottomrule
        
	\end{tabular}
	}
\end{table}

\subsection{Overall Performance}
\label{sec:overall_performance}
We compare our method with existing state-of-the-art approaches in Table~\ref{table:synthia} and Table~\ref{table:sim10k_kitti}, while the results evaluated by other metrics are provided in Table~\ref{table:sim10k_kitti_detail}.
We present two baselines: proposal-based Faster R-CNN~\cite{SRenNIPS15} and our fully-convolutional detector denoted as ``Ours (w/o adapt.)'', both without adaptation.
In all the tables, we denote global alignment and center-aware alignment as ``GA'' and ``CA'', respectively.
To understand how much domain gap our model reduces, we also present the ``Oracle'' results, in which the model is trained and tested on the target domain using our model.
Moreover, we consider two backbone architectures as our feature extractor: VGG-16 \cite{vgg} or ResNet-101 \cite{He_CVPR_2016}.
%

{\flushleft \bf{Weather Adaptation.}}
In Table~\ref{table:synthia}, we notice that our baseline without adaptation performs similarly (i.e., around 18$\%$) to the F-RCNN baseline, using the VGG-16 backbone. After adaptation, our method (GA + CA) improves our baseline by $17.6\%$ and performs the best compared to other methods in mAP$^r_{0.5}$, especially against the ones \cite{YChenCVPR18,ZHeICCV19,XZhuCVPR19} that adopt both global and instance-level alignments.
%
Overall, for both architectures, we consistently show that using the proposed center-aware alignment performs better than global alignment, and combining both is complementary and achieves the best performance.
%
%


{\flushleft \bf{Synthetic-to-real.}}
In the left part of Table~\ref{table:sim10k_kitti}, we show that our final model (GA+CA) using the VGG-16 backbone performs favorably against existing methods. 
%
We note that, compared to a recent method, SW-DA*~\cite{SaitoCVPR19}, that adds the augmented data into training via the pixel-level adaptation technique, our result is still better than theirs.
%
%
We also notice that the improvement from GA-only to GA+CA using the ResNet-101 backbone is not signiﬁcant. 
However, we will show that more performance gain can be achieved when using other mAP metrics with a higher standard later.
%


{\flushleft \bf{Cross-camera Adaptation.}}
In the right part of Table~\ref{table:sim10k_kitti}, we show that our method achieves favorable performance against others, and adding CA consistently improves the results, e.g., $8.8\%$ and $9.7\%$ gain compared to the baseline without adaptation, using VGG-16 or ResNet-101, respectively.


\begin{table*}[!t]
	\caption{
		More mAP metrics of adapting Sim10k/KITTI to Cityscapes using ResNet-101 as the backbone.
	}
	\label{table:sim10k_kitti_detail}
	\footnotesize
	\centering
    \resizebox{\textwidth}{!}{
	\begin{tabular}{lcccccc|cccccc}
		\toprule
		
		\multicolumn{1}{c}{} & \multicolumn{6}{c}{Sim10k $\rightarrow$ Cityscapes} & \multicolumn{6}{c}{KITTI $\rightarrow$ Cityscapes}\\
		
		\midrule
		 Method &  mAP &  mAP$^r_{0.5}$ &  mAP$^r_{0.75}$ &  mAP$^r_{S}$ &  mAP$^r_{M}$ &   mAP$^r_{L}$ &  mAP &  mAP$^r_{0.5}$ &  mAP$^r_{0.75}$ &  mAP$^r_{S}$ &  mAP$^r_{M}$ &   mAP$^r_{L}$ \\
		

        
        
        \midrule
        Ours (w/o adapt.)      & 23.1 & 41.8 & 22.4 & 5.1 & 26.8 & 46.6 & 15.9 & 35.3 & 12.8 & 1.5 & 17.8 & 36.5 \\
        Ours (GA)            & 26.4 & 50.6 & 25.2 & 5.7 & 26.3 & 57.3 & 18.8 & 42.3 & 14.7 & 5.0 & 24.5 & 35.9 \\
        Ours (CA)            & 26.8 & 51.1 & 26.3 & \textbf{7.5} & 27.9 & 54.6 & 20.3 & 43.6 & 17.3 & 4.1 & 25.4 & 40.8 \\

        Ours (GA+CA)         & \textbf{28.6} & \textbf{51.2} & \textbf{27.4} & 7.1 & \textbf{30.2} & \textbf{58.3} & \textbf{22.2} & \textbf{45.0} & \textbf{20.0} & \textbf{5.3} & \textbf{28.1} & \textbf{43.1} \\
        
        \greyrule
        Oracle (ResNet-101)              & 44.6 & 70.4 & 46.2 & 15.7 & 49.2 & 79.2 & 44.6 & 70.4 & 46.2 & 15.7 & 49.2 & 79.2 \\

        \bottomrule
        
	\end{tabular}
	}
\end{table*}


{\flushleft \bf{More Discussions.}}
Although the CA-only model performs competitively against the GA-only model, they essentially focus on different tasks.
For global alignment, it tries to align image-level distributions, which is necessary to help reduce the domain gap but may focus too much on background pixels.
For our center-aware alignment, we focus more on pixels that are likely to be the foreground, in which the alignment process considers foreground distributions more.
As such, they act as a different role, in which combing both is complementary to further improve the performance (i.e., GA+CA).

In addition, in Table \ref{table:synthia}, we notice that the performance of some categories that are underrepresented such as \textit{truck} and \textit{mbike} is lower than that of other categories. One reason is that these categories contain less foreground pixels in the source domain, in which our center-aware alignment may pay less attention to them. One could adopt a stronger backbone (e.g., ResNet-101 in Table \ref{table:synthia}) to improve the performance or use the category prior that allows the model to focus more on those underrepresented categories, which is not in the scope of this work and could be one future work.

\subsection{More Results and Analysis}
In this section, we provide detailed analysis in the proposed method with more mAP measurements. In addition, we visualize our center-aware maps and more results are provided in the supplementary material.

{\flushleft \bf{More mAP Metrics.}}
In Table~\ref{table:sim10k_kitti_detail}, we show more mAP metrics than mAP$^r_{0.5}$, to analyze where our method helps the detector adapting to different scenarios.
%
%
On the Sim10k case, as discussed in Section \ref{sec:overall_performance}, we observe that our full model using ResNet-101 does not improve mAP$^r_{0.5}$ a lot compared with the GA-only model. However, we show that under a more challenging case, e.g., mAP$^r_{0.75}$, mAP$^r_{S}$ and mAP$^r_{M}$, adding CA improves results over GA-only by $2.2\%$, $1.4\%$, and $3.9\%$, respectively.
It validates the usefulness of our center-aware alignment for challenging adaptation cases.
Similar observations could be found in the KITTI case.
Such measurements also suggest an interesting aspect for domain adaptive object detection to better understand its challenges.

{\flushleft \bf{Multi-scale Alignment.}}
To verify the effectiveness of our multi-scale alignment scheme, we conduct an ablation study on Sim10k $\rightarrow$ Cityscapes using the ResNet-101.
In Table \ref{table:multi_layer}, we compare results using all the scales ($F^3 \sim F^7$), three scales ($F^5 \sim F^7$) via removing the bottom two scales, three scales ($F^3 \sim F^5$) via removing the top two scales, and a single scale $F^5$.
Note that, we choose the single scale as $F^5$ since it is the middle scale, which has the most influential impact.
We show that adding more scales gradually improves the performance on all the metrics, which validates the usefulness of our proposed multi-scale alignment method.
Moreover, $F^3 \sim F^5$ is responsible for smaller objects, in which the mAP$^r_{S}$/mAP$^r_{M}$ results are better than the $F^5 \sim F^7$ ones.
In contrast, mAP$^r_{L}$ is better for $F^5 \sim F^7$ as it handles larger objects.
%
This indicate that our multi-scale alignment is effective for handling various size of objects.


\begin{figure*}[t]
	\centering
    \resizebox{0.9\textwidth}{!}{
	\begin{tabular}
		{@{\hspace{0mm}}c@{\hspace{1mm}} @{\hspace{0mm}}c@{\hspace{1mm}} @{\hspace{0mm}}c@{\hspace{1mm}}
			@{\hspace{0mm}}c@{\hspace{1mm}} @{\hspace{0mm}}c@{\hspace{0mm}}
		}
		\includegraphics[width=0.22\linewidth]{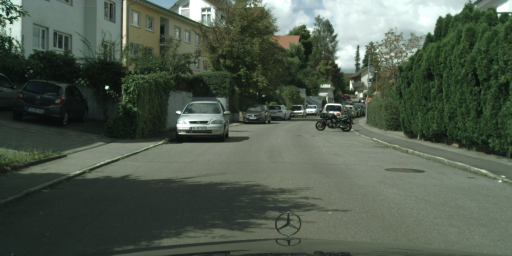} &
		\includegraphics[width=0.22\linewidth]{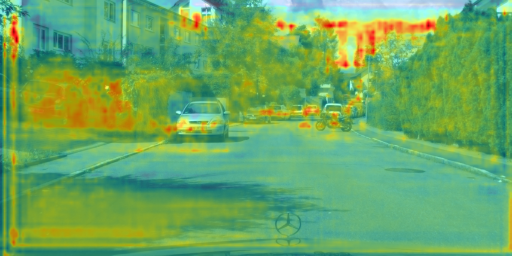} &
		\includegraphics[width=0.22\linewidth]{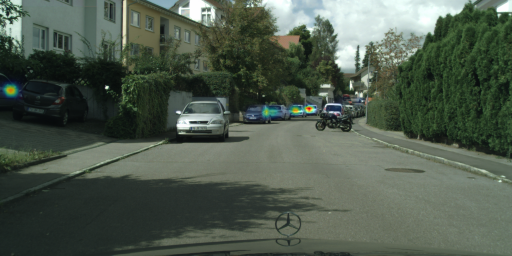} &
		\includegraphics[width=0.22\linewidth]{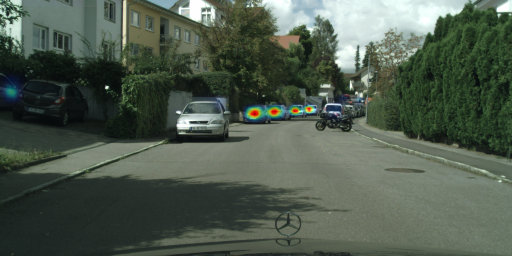} \\
		
		\includegraphics[width=0.22\linewidth]{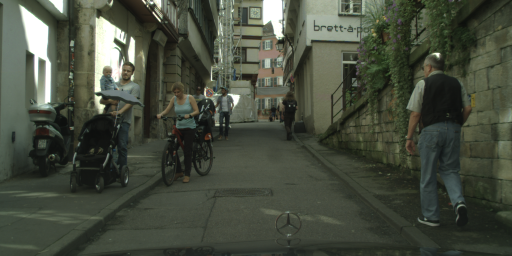} &
		\includegraphics[width=0.22\linewidth]{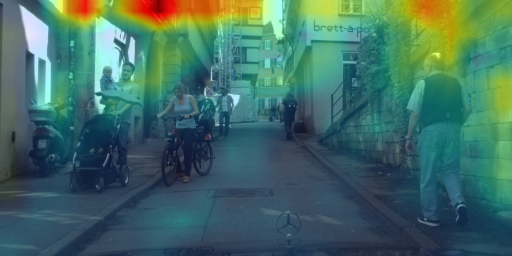} &
		\includegraphics[width=0.22\linewidth]{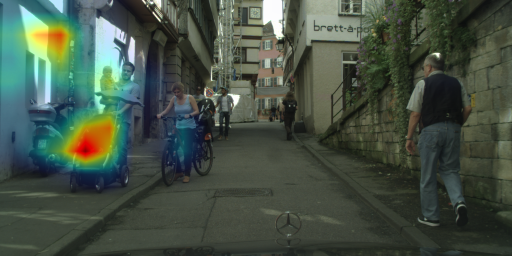} &
		\includegraphics[width=0.22\linewidth]{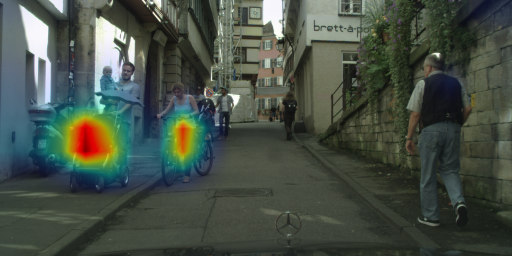} \\
		
		Target Image & w/o Adaptation & GA & GA + CA \\
		
	\end{tabular}
	}
	\caption{
	    \textbf{Comparisons of response maps on Sim10k-to-Cityscapes.}
		The maps on the first row are extracted from the feature layer $F^3$ which focuses on smaller objects, while the second row is for the feature layer $F^6$. After adding the proposed center-aware alignment, the model could focus more on the objects and reduce background noises.
	}
	\label{fig: vis_GA_CA}
\end{figure*}

\begin{table} [t]
	\caption{
		\textbf{Ablation study of the proposed multi-scale alignment.}
		We adopt ResNet-101 as the backbone in this study.
	}
	\label{table:multi_layer}
	\small
	\centering
    \resizebox{0.6\textwidth}{!}{
	\begin{tabular}{lcccccc}
		\toprule
		
		\multicolumn{7}{c}{Sim10k $\rightarrow$ Cityscapes} \\
		\midrule
		
		Aligned Scale &  mAP &  mAP$^r_{0.5}$ &  mAP$^r_{0.75}$ &  mAP$^r_{S}$ &  mAP$^r_{M}$ &   mAP$^r_{L}$ \\
		
		\midrule
        w/o adapt.        & 23.1 & 41.8 & 22.4 & 5.1 & 26.8 & 46.6 \\
        $F^5$              & 24.2 & 48.9 & 22.4 & 5.7 & 24.0 & 52.4 \\
        $F^3 \sim F^5$      & 26.2 & 48.7 & 25.0 & 6.9 & 28.7 & 53.0 \\
        $F^5 \sim F^7$      & 26.1 & 49.2 & 25.8 & 6.2 & 26.8 & 54.8 \\
        $F^3 \sim F^7$      & \textbf{28.6} & \textbf{51.2} & \textbf{27.4} & \textbf{7.1} & \textbf{30.2} & \textbf{58.3} \\
        
		\bottomrule
	\end{tabular}
	}
\end{table}

{\flushleft \bf{Qualitative Analysis.}}
We first show some example results of the response map that our method tries to localize the object.
In Fig. \ref{fig: vis_GA_CA}, the baseline without adaptation has difficulty to find any object centers, while our global alignment method is able to localize some objects.
Adding the proposed center-aware alignment enables our method to discover more object centers at different object scales.
We also note that, each scale in our model may focus on a different size of object, e.g., the upper example in Fig. \ref{fig: vis_GA_CA} may miss larger objects. However, those objects missing at a smaller scale could be identified at another scale.
%

%

\section{Conclusions}
In this paper, we propose a center-aware feature alignment method to tackle the task of domain adaptive object detection.
Specifically, we propose to generates pixel-wise maps for localizing object regions, and then use them as the guidance for feature alignment.
To this end, we develop a method to discover center-aware regions and perform the alignment procedure via adversarial learning that allows the discriminator to focus on features coming from the object region.
In addition, we design the multi-scale feature alignment scheme to handle different object sizes.
Finally, we show that incorporating global and center-aware alignments improves domain adaptation for object detection and achieves state-of-the-art performance on numerous benchmark datasets and settings.

\noindent
\textbf{Acknowledgment.}
This work was supported in part by the Ministry of Science and Technology (MOST) under grants MOST 107-2628-E-009-007-MY3, MOST 109-2634-F-007-013, and MOST 109-2221-E-009-113-MY3, and by Qualcomm through a Taiwan University Research Collaboration Project.

\clearpage
%
%
\bibliographystyle{splncs04}
\bibliography{egbib}
\end{document}